# Regulating autonomous and agentic AI

**Chris Reed, Alex Austria, Anmol Bharuka, Pragnitha Mandava, Khushiya Mujawar, Luka Shakhkulashvili***

**Abstract**

Regulating activities where regulatees use autonomous and agentic AI is challenging. Regulatory assumptions about regulatee knowledge and control no longer hold true; much of that lies elsewhere in the AI supply chain which thus needs to be brought within the scope of regulation. Governance systems for autonomous AI cannot replicate existing governance models, but need a fresh approach. Retrospective supervisory oversight becomes ineffective as a risk management tool, and AI autonomy generates new systemic risks which require new solutions.

This paper investigate four regulatory systems: UK regulation of content platforms, data protection, UK financial services, and the EU AI Act's cross-sectoral regime. It analyses the challenges posed by autonomous and agentic AI and proposes potential solutions which regulators might adopt. These will transform regulation from a reactive process to an active one, and assist it in adapting to the challenges of AI autonomy.

*All URLs cited were last visited 22 July 2026*

## 1 Introduction

Regulation aims to ensure that particular activities are undertaken to appropriate standards. It does so by prescribing how those it regulates, the regulatees, should act when carrying out those activities. Regulated organisations are responsible for compliance, and act through their employees who are responsible to the employer (and sometimes also directly to the regulator) for complying with the regulation's demands.

This responsibility is unproblematic where the individuals concerned control the processes, and in particular the technologies, they use to carry out the regulated activity. Those who have control can be held responsible for what they control.

However, artificial intelligence (AI) technologies can break the link between control and responsibility if they act autonomously, making decisions and implementing them without human intervention. It is now the AI which controls the performance of the regulated activity, while current regulatory approaches have been devised on the basis that humans are exercising control.

We thus need to ask whether those regulatory approaches still work to achieve their regulatory objectives, and if not, what new approaches might be devised to deal with autonomous AI. There is clearly no regulatory difficulty if the AI does what its user wanted and expected it to do. So the issue

* Chris Reed is Professor of Electronic Commerce Law at the Centre for Commercial Law Studies, Queen Mary University of London. Alex Austria, Anmol Bharuka, Pragnitha Mandava, Khushiya Mujawar, Luka Shakhkulashvili are all postgraduate student researchers for the Cloud Legal Project.
This research was supported by the Cloud Legal Project at the Centre for Commercial Law Studies, Queen Mary University of London, and we are grateful to Microsoft for the generous financial support that has made this project possible. Responsibility for views expressed remains with the authors.

arises where the AI's autonomous actions were undesired and unexpected, and would amount to regulatory breaches if those actions were undertaken by a human. This raises two questions:

- To what extent, and on what basis, does current regulation make AI users responsible for the autonomous actions of an AI?
- How *ought* regulation to deal with this issue?

This paper attempts answers by investigating three regulatory regimes aimed at users of AI technology – online platforms in respect of the content they make available, data protection, and financial services. The paper also considers the EU's AI Act[1] which is aimed at the developers and deployers of AI technology who supply regulatees, and also at those regulatees who put AI systems to use.

## 1.1 AI autonomy

AI autonomy does not alter regulatory responsibility where the AI's role is merely to analyse and present information. In those cases a human, in theory at least, makes the final decision to act as recommended by the AI's outputs. In practice there is a clear risk that those humans will at some point cease to exercise independent judgment and just rubber-stamp the AI's suggested actions,[2] but it remains the responsibility of regulatees to counter this danger via staff training and organisational processes.

Full autonomy is where the AI can make decisions and also put them into effect without human approval. We find such autonomy in AIs which operate control systems for physical processes; an obvious example would be self-driving vehicles. Where these are part of a regulated activity, the form of regulation tends to focus on testing and certification before the technology can be used.[3] This type of regulation is specific to the activity in question so that general conclusions are hard to draw.

By contrast, agentic AI is a general purpose technology which can be customised for individual applications. Agentic AI builds on the Large Language Models (LLMs) which power generative AI, with three additional features:[4]

---

[1] Regulation (EU) 2024/1689 of the European Parliament and of the Council of 13 June 2024 laying down harmonised rules on artificial intelligence (Artificial Intelligence Act) [2024] OJ L (the AI Act). Unhelpfully, the AI Act defines the developer of an AI systems as its 'provider' (art 3(3)) and its 'deployer' (art 3(4)) is what we term the 'user' in this paper. We think it important to identify the entities who are intermediate between developer and user, because they modify the AI for a more specific use, and the term 'deployer' is most appropriate for this purpose. For a more detailed explanation of the roles of developer, deployer and user, see Philipp Hacker, Andreas Engel and Marco Mauer, 'Regulating ChatGPT and Other Large Generative AI Models' in *Proceedings of the 2023 ACM Conference on Fairness, Accountability, and Transparency (ACM 2023)* 1112, 1115-16.

[2] Alisa Küper and Nicole Krämer, 'Psychological traits and appropriate reliance: Factors shaping trust in AI' (2025) 41 *International Journal of Human–Computer Interaction* 4115; Artur Klingbeil, Cassandra Grützner and Philipp Schreck, 'Trust and reliance on AI—An experimental study on the extent and costs of overreliance on AI' (2024) 160 *Computers in Human Behavior* 108352.

[3] See eg the UK Automated Vehicles Act 2024, which establishes schemes for the certification of autonomous vehicles and their operators and includes provision for what is effectively a driving test for those vehicles.

[4] Plaat et al summarise the abilities of agentic AI as: '(1) *reasoning* capabilities, (2) an interface to the outside world in order to *act*, and (3) a social environment with other agents with which to *interact*.' Aske Plaat, Max van Duijn, Niki van Stein, Mike Preuss, Peter van der Putten, and Kees Joost Batenburg, 'Agentic large language models, a survey' (2025) *arXiv preprint arXiv:2503.23037*, 8.

- The AI user sets its goal, but the AI itself determines the means to be used to achieve that goal rather than following a process set by its user.[5] The AI also has power to adapt those means as the information it possesses is updated.[6]
- The AI can recruit other AIs as assistants and use information tools (tool-invocation capability), in effect assembling and running an expert team to achieve its goals.
- The AI acts directly to achieve the goal, interacting with external systems to initiate transactions, make payments, and so-on.

The important distinction here is between goal autonomy, which only the human user possesses, and solution autonomy, which is granted to the agentic AI.[7] Because of that solution autonomy the actions undertaken by an agentic AI system in pursuit of its goal can be unexpected and unpredictable. This produces what Nannini and others term as 'runtime behavioural drift,' where the agent's operational profile shifts through accumulated interaction.[8] Uncertainty here is made worse because the AI might recruit untrustworthy sub-agents,[9] prioritise the user's aims over the interests of others such as customers,[10] or even autonomously break free of any constraints placed on its actions if they hinder it from achieving its goal.[11]

If this were all, regulators might simply analogise agentic AI with the staff employed by regulatees, for whose actions the regulatee has always had responsibility. But agentic AI is more unpredictable than employees because its decision-making does not follow the same 'mental' processes as a human decision-maker. As just two examples, agentic AI *mimics* causal reasoning rather than employing it and so may confuse correlation and causation leading to undesired actions,[12] and an externally-developed agentic AI may have a very different tolerance for particular risks than its user, differences which are not easily identifiable or remediable.[13] Additionally, reliable information about how and why the AI reached its decision is not available.[14]

---

[5] An agentic AI has autonomous planning capacity, the ability to break down high-level objectives into sub-tasks without human prompting or intervention at each step.

[6] See Erik Miehling, Karthikeyan Natesan Ramamurthy, Kush R.Varshney, Matthew Riemer, Djallel Bouneffouf, John T Richards, Amit Dhurandhar et al, 'Agentic Ai needs a systems theory' (2025) *arXiv preprint arXiv:2503.00237*, 4, arguing that 'the system [needs to be] able to modify its behavior when the relationship between actions and outcomes changes. Without adaptation, the system would be unable to maintain goal-directed behavior over time.'

[7] Ibid, 4.

[8] Luca Nannini et al, 'AI Agents Under EU Law: A Compliance Architecture for AI Providers' (2026) *arXiv preprint arXiv:2604.04604v1*, 15.

[9] Miehling et al, n 6, 8.

[10] Alan Chan, Rebecca Salganik, Alva Markelius Chris Pang, Nitarshan Rajkumar, Dmitrii Krasheninnikov, Lauro Langosco et al, 'Harms from increasingly agentic algorithmic systems' (2023) *Proceedings of the 2023 ACM Conference on Fairness, Accountability, and Transparency* 651, 657.

[11] Miehling et al, n 6, 1-2. This might include behaviour which would be criminal if undertaken by a human – see 'OpenAI: Rogue AI agent hacked rival platform by itself', *The Times* 22 July 2026.

[12] Miehling et al, n 6, 5.

[13] Hayley Clatterbuck, Clinton Castro and Arvo Muñoz Morán, 'Risk alignment in agentic AI systems' (2024) *arXiv preprint arXiv:2410.01927*, 6-21.

[14] An LLM-based AI can be asked to explain its reasoning and will produce a plausible explanation, a 'chain-of-thought'. However, that explanation has no necessary connection with the truth. Fazl Barez et al, 'Chain-of-thought is not explainability' *Preprint, alphaXiv* (2025): v1, https://aigi.ox.ac.uk/wp-content/uploads/2025/07/Cot_Is_Not_Explainability-1.pdf

## 1.2 The AI supply chain

Producing, deploying, and actually using an AI system typically involves a chain of different interdependent actors, tied together by flows of data between them.[15] This supply chain is driven by the heavy processing needs of generative AI: major cloud providers supply pre-built, generalised AI capabilities that other organisations take and incorporate into their own products and services, which are often specialised for particular sectors of activity.[16]

At one end of the chain sits the developer, the organisation that creates and trains the model in the first place. That trained model is then taken up by the deployer, who adapts it for a specific use case, typically on cloud infrastructure. There may be several deployers in a single chain, working in sequence or in parallel, in a structure that Hacker et al have identified as comparable to OEM value chains in manufacturing.[17] From there, the model reaches the user, who generates actual output from it. Users may be professional, meaning firms or institutions using the model in the course of their business, or non-professional. In regulated sectors, the professional user will typically be a firm with its own compliance obligations, such as a bank or investment firm. At the far end of the chain is the recipient, the person or organisation that experiences the AI's action without having had any part in producing it.

This chain structure complicates the central regulatory question: who should be responsible for what, and where should liability fall when something goes wrong. Developers have knowledge about the limitations and known failure modes of their models, but do not know how those models will actually be used. Deployers are often smaller and less technically capable than developers, yet they are the ones who customise the model for use in the contexts where harm actually occurs. Users have the fullest understanding of how they will use the AI, and thus the risks if it fails to work as they expect, but may be in ignorance of limitations and failure modes identified further back in the supply chain.

This means that no single actor in the chain has a complete picture of how the whole system works. Moral responsibility is distributed across actors who may not be easy to identify, and no single entity controls the chain from one end to the other.[18] Thus placing the full burden of regulatory obligations on any one party alone fails to match where moral responsibility lies, and makes them liable for matters which are outside their own knowledge and control.[19]

# 2 Autonomous AI problems for regulators

## 2.1 Knowledge and control

Regulation allocates responsibility to regulated entities on the assumption that the risks their activities create can be predicted, and thus that measures to control those risks can be implemented. It also assumes that the regulatee will be able to explain, and thus be accountable for, any failures of control. It is the humans working for the regulated entity who have to exercise foresight and devise controls, so responsibility and accountability devolves to those humans. This is explicit in the UK's financial regulation[20] and implicit in the other regulatory systems examined.

[15] Jennifer Cobbe, Michael Veale and Jatinder Singh, 'Understanding Accountability in Algorithmic Supply Chains' (2023) Proceedings of the 2023 ACM Conference on Fairness, Accountability, and Transparency 1186.
[16] ibid, 1189-90.
[17] Hacker, Engel and Mauer, n 1.
[18] Cobbe, Veale and Singh, n 15, 1192.
[19] Hacker, Engel and Mauer, n 1, 1116.

Regulation also assumes that those humans both know the risks created by their activity and are able to control them. Fraser and Suzor argue that coherent accountability requires the responsible entity to possess both knowledge and the ability to control at the same time.[21] AI autonomy weakens the ability of humans here. Selbst proposes that many, perhaps most, AI errors fall outside the realm of foreseeability unless the AI is explainable or transparent,[22] so that the necessary knowledge of risk is absent. The unpredictable conduct of autonomous AI makes control more difficult, perhaps even impossible, thus breaking the causal link which enables harm to be attributed to a specific human. Motagnani, Najjar and Davola argue that this produces what they describe as an 'accountability gap'.[23]

The UK FCA's Mills Review recognises that this is an issue which regulators will need to resolve. Delegation of functions to autonomous and agentic AI creates 'a potential gap between the outcomes of AI mediated decisions and the ability of the regulator to identify a de facto responsible party or for responsible individuals to exercise "meaningful human control" where appropriate.'[24]

One possible response might be to assign responsibility to humans on the basis that the AI is the mechanism they have chosen to undertake the activity. This might be analogised to appointing a human agent,[25] or the use of some machine or tool,[26] either of which might fail to work as intended. However, legal responsibility for the actions of agents or the failure of machines is not absolute, but rather depends on foreseeability and the precautions taken to prevent those foreseeable risks.[27]

This problem of knowledge and control does not mean that regulation is unable to hold regulated entities responsible and accountable for the actions of their autonomous AI systems. But it does suggest that doing so on the basis that humans can foresee risks and implement control measures may become increasingly unworkable.

---

[20] Via the UK FCA Senior Managers and Certification Regime, https://www.fca.org.uk/firms/senior-managers-certification-regime.

[21] Henry L Fraser and Nicolas P Suzor, 'Locating fault for AI harms: a systems theory of foreseeability, reasonable care and causal responsibility in the AI value chain' (2025) 17(1) *Law, Innovation and Technology* 103, 106.

[22] AD Selbst, 'Negligence and AI's Human Users' (2020) 100(4) *Boston University Law Review* 1362.

[23] Maria Lillà Montagnani, Marie-Claire Najjar and Antonio Davola, 'The EU Regulatory Approach(es) to AI Liability, and its Application to the Financial Services Market' (2024) Computer Law & Security Review 14–15.

[24] UK FCA, Review into the long-term impact of AI on retail financial services (The Mills Review) (2026), 81, https://www.fca.org.uk/publications/calls-input/review-long-term-impact-ai-retail-financial-services-mills-review.

[25] See eg UK Jurisdiction Taskforce, Consultation on the Draft Legal Statement on Liability for AI Harms under the Private Law of England and Wales (UKJT Consultation Paper, January 2026); see also the annexed Draft Legal Statement on Liability for AI Harms under the Private Law of England and Wales, 49 – 51, https://lawtechuk.io/ukjt/public-consultation-liability-for-ai-harms-under-the-private-law-of-england-and-wales/; Chris Reed, 'Autonomy, Responsibility and Agentic AI' (forthcoming, *Journal of Business Law*) https://papers.ssrn.com/sol3/papers.cfm?abstract_id=5598471.

[26] See Xiaoshui Zhai, 'Autonomous systems: how should contract law treat such autonomy? (2026) *Information & Communications Technology Law*, 1, 7 and 27, noting that the UNCITRAL, 'Model Law on Automated Contracting', https://uncitral.un.org/sites/uncitral.un.org/files/mlac_en.pdf, makes the user of an automated contracting system responsible for transactions unless the other party know or ought to have realised that the system was acting in an unintended way.

[27] Reed, n 25.

## 2.2 Accountability in the supply chain

AI supply chains disrupt the logic of AI user responsibility. In a multi-party deployment, the developer builds the model and determines its foundational architecture, the deployer integrates the model into a service, and downstream operators may layer additional processing on top. All this happens before the AI system is made available to the regulatee.

The ways in which those developers, deployers and downstream operators perform their activities build risks into the final AI system.[28] The regulated user is supposed to manage those risks but may not even be aware of them for three reasons:

- Knowledge of risks is siloed within supply chain members, most likely for reasons of commercial confidentiality, and thus is not passed down the supply chain to the user;
- The supply chain member may not even be aware that they have created a risk, for the reasons discussed in section 2.1; or
- The risk may arise from the interaction of the technologies developed or integrated by supply chain members rather than from one member's isolated decision,[29] and thus not manifest itself until the AI is put into use by the regulated user.

The result is accountability fragmentation.[30] Where multiple actors contribute to a processing outcome, allocating responsibility for the whole to a single member becomes doctrinally incoherent.[31]

Arguably, actors who contribute to the creation or increase of risk should also be part of the accountability analysis. The point is not to reduce regulated user responsibility, but to allocate responsibility according to each actor's contribution, knowledge, control, and preventive capacity. The regulatee would remain the primary responsible actor, but regulation should better recognise where developers, deployers, and integrators also shape the relevant risk.[32] We discuss this idea further in section 4.

## 2.3 Risk management through governance

As we will see in section 3.4, the primary compliance mechanism required by regulation is the implementation of appropriate governance by the regulated entity. The aims of that governance are to identify risks and implement controls to prevent or minimise their occurrence. Thus the issues of lack of foreseeability about how an agentic AI will act, and the unpredictability of those actions which makes control difficult, also arise here. Additionally, the distribution of practical (though not necessarily regulatory) responsibility across the AI supply chain (see sections 1.2 and 2.2) generates uncertainty about where governance should be undertaken, and by whom.

[28] Hacker, Engel and Mauer, n 1, 1115–16.
[29] Teresa Rodríguez de Las Heras Ballell, 'Mapping Generative AI Rules and Liability Scenarios in the AI Act, and in the Proposed EU Liability Rules for AI Liability' (2025) 1 *Cambridge Forum on AI: Law and Governance* e5, 7–8, 20.
[30] Cobbe, Veale and Singh, n 15, 1194.
[31] Jennifer Cobbe and Jatinder Singh, 'Artificial Intelligence as a Service: Legal Responsibilities, Liabilities, and Policy Challenges' (2021) 42 Computer Law & Security Review 105573, 7; Dimitra Kamarinou, Christopher Millard and Jatinder Singh, 'Machine learning with personal data: profiling, decisions and the EU general data protection regulation', in *29th Conference on Neural Information Processing Systems (NIPS 2016)*.
[32] Anirban Mukherjee and Hannah Hanwen Chang, 'Operational Agency: A Permeable Legal Fiction for Tracing Culpability in AI Systems' *(forthcoming 2026) SMU Science and Technology Law Review*, 50 https://ssrn.com/abstract=5680063; Fraser and Suzor, n 21, 135–36.

The regulatory systems investigated for this paper all take similar approaches to technology governance, identifying four main stages:

- Risk identification
- Implementation of safeguards
- Monitoring for failure
- Improvement of safeguards and correction of systems failure

Governance systems are the primary line of defence, but can only work if they generate the knowledge needed to identify and mitigate risk. As explained in the previous sections, AI autonomy obscures knowledge and hampers control, which means that human governance systems are at risk of failing the first two stages.

Agentic AI also challenges the third and fourth stages. Current regulatory frameworks assume that regulatees can supervise and intervene in automated systems, but to do this they need knowledge about how and why the system is acting. Oversight is meaningful only where reviewers have visibility, expertise, authority, and practical capacity to act, but agentic AI outputs are dynamic and system behaviour may be difficult to observe or explain. Agentic AI's inability to explain itself reliably makes correction challenging. The inevitable lack of insight into the AI's behaviour creates the risk that governance becomes a performative exercise, formal compliance rather than a mechanism for substantive control.[33]

Additionally, regulation can disincentivise governance improvement because better governance generates more detailed evidence to justify imposing regulatory liability. Minimal monitoring permits plausible ignorance claims, while rigorous monitoring eliminates them and triggers intervention obligations .[34] When knowledge arrives too late to prevent harm but early enough to establish responsibility,[35] regulatees will naturally prefer ignorance.

## 2.4 Supervision problems

Regulators need the ability to supervise how regulatees use AI, but the challenges from autonomous AI systems identified in sections 2.1-2.3 make effective supervision difficult. Regulators need to understand how the AI *might* act, in order to decide whether the regulatee has properly considered and guarded against the potential risks before putting it in to operation. Regulators also need to be able to understand the reasons for any failures through analysis after the event, so as to evaluate how the regulatee proposes to prevent recurrence.

First, the nature of generative AI as a technology for predicting likely human responses to prompts obscures knowledge and hampers control. A generative AI is likely to produce a plausible response which is within the range of responses of a human actor, but is not certain to do so,[36] and its precise response is unpredictable and irreproducible. If the AI has autonomous capabilities to implement its decisions, this means that its actions are hard to predict, and equally difficult to explain after the event.

[33]Anirban Mukherjee and Hannah Hanwen Chang, 'Agentic AI: Autonomy, Accountability, and the Algorithmic Society' (2025) 9–12, 19-22, arXiv preprint arXiv:2502; Muhammad Ajmal and Azmat Islam, 'Who Leads When AI Acts? Authority and Accountability in Agentic Organizations' (2025) 3(12) *Policy Journal of Social Science Review* 629, 631, 640.
[34]Information Commissioner's Office, 'What is the impact of Article 22 of the UK GDPR on fairness?' (2025) https://ico.org.uk/for-organisations/uk-gdpr-guidance-and-resources/artificial-intelligence/guidance-on-ai-and-data-protection/how-do-we-ensure-fairness-in-ai/what-is-the-impact-of-article-22-of-the-uk-gdpr-on-fairness/.
[35]European Parliament Research Service, Andrea Bertolini and others, 'Artificial Intelligence and Civil Liability: A European Perspective' (Study for JURI Committee, January 2025) PE 776.426.

Second, one aim of regulatory supervision is to identify patterns of failure which regulatees can then be required to guard against. The dynamic nature of agentic AI decisions means that although individual breaches can be identified, they offer little insight into whether that AI service, or similar services, are likely to act the same way again. Generalising the nature and reasons for breaches to generate insights for industry level adoption or regulatory guidance is thus challenging.

The obvious solution is to institute monitoring of AI actions, either by the regulatee or even directly by the regulator, coupled with guardrails to prevent the AI from implementing undesired actions. This solution is also challenged by AI autonomy. Reinforcement learning training may incentivise AI agents to develop goal-directed strategies to circumvent monitoring and guardrails, prioritising achieving the AI's goal over regulatory compliance.[37]

## 2.5 Systemic risks

In addition to protecting individuals and customers of regulatees, some regulators have formal responsibility for controlling risks to the wider system within which their regulatees act. This is the case for the UK's FCA, but it is likely that OFCOM and the ICO also take into account how their regulation influences the totality of the systems within which regulatees operate.

Agentic AI introduces novel systemic risks, which we explain below. Our main focus is financial sector regulation, because of its clear focus on maintaining stability in the financial system, but we also suggest how these risks might manifest in other sectors.

The first new risk is correlated behaviour by agents. Correlated behaviour by humans is always intentional, and is deterred by the demands of competition with others and by competition law. These constraints do not apply to AI agents - where multiple firms deploy agents trained on overlapping data and similar foundation models, those agents may converge on similar strategies for achieving their goals without collaborating. This can produce herding effects which are stronger and emerge more rapidly than previously and can lead to rapid market movements which threaten the system.[38] We can foresee similar risks of correlation among AI agents deciding when personal data should be collected or shared, and of AI content monitoring systems converging on an incorrect view about, for example, the types of content which should not be constrained on free speech grounds..

The second is infrastructure concentration, with critical functions all depending on a small number of cloud and model providers which thus creates a single point-of-failure risk. In the financial sector it is already recognise that this risk is inadequately dealt with by traditional outsourcing rules.[39] The

[36] For example, there is the possibility that it might hallucinate information to fill gaps in the data on which it was trained - Yujie Sun, Dongfang Sheng, Zihan Zhou, and Yifei Wu, 'AI hallucination: towards a comprehensive classification of distorted information in artificial intelligence-generated content' (2024) 11 *Humanities and Social Sciences Communications* 1.

[37] Nannini et al, n 8, 13–14.

[38] House of Commons Treasury Committee, Artificial Intelligence in Financial Services (Fifteenth Report of Session 2024–26, HC 684, 2026), https://committees.parliament.uk/publications/51128/documents/283671/default/, 6.

[39] Ibid, 8 (identifying over-reliance on a small number of US-based technology firms as a major sectoral vulnerability).

EU's DORA regulation partially addresses this by treating ICT third parties as critical infrastructure,[40] and the UK Treasury Committee is expected to designate major AI providers as Critical Third Parties (CTP) under the analogous CTP regime[41] by the end of 2026. This will bring designated providers partially within the UK's financial sector regulation, primarily by imposing similar governance obligations to regulated firms.[42] Similar issues arise for any data controller which uses third parties or agentic AI to process personal data, but the existing mechanisms for managing this risk are much more limited in scope, particularly cross-border.[43]

More generally, all systems are put at risk by coordination failures by their participants. The financial system has devised macroprudential instruments to manage these risks, but they focus on human failure, not failures arising from emergent inter-agent interactions. For this reason the UK Treasury Committee has accordingly recommended AI-specific stress testing by the end of 2026.[44] The risks of this kind to other systems are less clear, but may also need regulatory intervention in some cases.

# 3 Regulated activities

The three systems we examine regulate different *activities*. The core activities are: making available online content provided by third parties; processing personal data; and providing specified financial services. Additionally, most financial services regulation requires the regulated entity to be authorised by the regulator before providing services.[45] The aim is to prevent these activities from being carried out in ways which cause specific harms: societal and individual harm from content shared online; harm to data subjects from the unfair processing of their personal data; and harm to customers from inadequate financial advice or services as well as harm to the financial system as a whole.

The EU AI Act applies across all sectors, and regulates three activities: making an AI system available, placing it on the market, and putting it into service. Its primary aims are to ensure in advance that AI systems are safe in the broadest sense,[46] and that users and supply chain members have systems in place to detect and correct safety failures.

Regulation works to achieve its aims in two ways. First, it specifies how the activity in question should be undertake (or how it should not be undertaken), in other words it tries to control or influence the behaviour of regulatees. Second, it imposes sanctions on regulatees who fail to avoid causing or contributing to harms which are to be prevented. AI autonomy challenges both these approaches for the reasons explained in section 2. In particular, the problems of knowledge and control make it difficult to specify how regulatees should act when implementing and operating agentic AI systems, and they also make it harder to justify imposing responsibility and sanctions where the harm was caused by an AI's autonomous acts.

---

[40] Regulation (EU) 2022/2554 of the European Parliament and of the Council on digital operational resilience for the financial sector [2022] OJ L333/1 (DORA) recital 20.
[41] See section 4.1.2.
[42] The regulatory obligations are set out in Bank of England Supervisory Statement SS6/24, Operational resilience: Critical third parties to the UK financial sector, https://www.bankofengland.co.uk/-/media/boe/files/prudential-regulation/supervisory-statement/2024/ss624-november-2024.pdf.
[43] Bjørn Aslak Juliussen, Elisavet Kozyri, Dag Johansen and Jon Petter Rui, 'The third country problem under the GDPR: enhancing protection of data transfers with technology' (2023) 13 *International Data Privacy Law* 225.
[44] Treasury Committee (n 1) 6 (recommending AI-specific stress testing by the end of 2026).
[45] See eg UK Financial Services and Markets Act 2000 s 19, which prohibits any person from carrying out the specified activities (s 22) unless they are authorised or exempt.
[46] The AI Act does not state this explicitly, but its origins are in the EU's product safety regime and its main focus is on risk prevention.

## 3.1 Outcome-based v care and skill obligations

Responsibility, leading to liability, is greatly affected by the nature of the regulatory obligation in question. Data protection and UK financial services regulation are both outcome-based regimes. This means that the regulatee is required to achieve a specified set of outcomes[47] and, in theory at least, the reason for any failure is irrelevant. By contrast, platform content control regulation requires the regulatee to exercise care and skill to identify and mitigate risks,[48] primarily via appropriate governance systems (see section 3.4). It might be thought that outcome-based regulation avoids all the problems of AI autonomy – either the outcome has been achieved, or it has not.

The outcome obligations of the GDPR are ostensibly absolute. Article 5(2) makes the data controller not only responsible for achieving them, but the controller is also required actively to demonstrate compliance. The controller makes decisions about the purpose and means of processing, and if the controller decides to implement an agentic AI system which can autonomously determine those matters, the AI's actions cannot be separated from the decision-making process conducted by the controller when deploying the software. The decision to use the AI is a decision about the purposes and means of processing. Thus, there is no basis for arguing that the controller is not liable for the autonomous actions of the AI.

This is certainly the position taken by data protection authorities (DPAs). In March 2023 the Italian data protection authority (DPA), *Garante per la protezione dei dati personali*, introduced a temporary prohibition on using OpenAI's ChatGPT service because ChatGPT was autonomously deciding to include personal data in its outputs. Those AI decisions were not based on any legitimate ground and took no account of the transparency and purpose limitation obligations of the GDPR.[49] A little earlier the French, UK, Italian and Greek DPAs took coordinated enforcement action against Clearview AI, holding that the autonomous scraping and indexing of over three billion facial images was the controller's responsibility, not an autonomous act that could be attributed to the algorithm itself.[50]

AI autonomy as an affirmative defence was raised by the Dutch tax authority, the Belastingdienst, when it implemented an automated scoring system which marked thousands of benefit applicants from certain minorities as suspected of fraud and led to their debts being recovered in a way that caused serious long-term damage. This practice constitutes unlawful discriminatory profiling, but the tax authority argued that it was not in breach of this obligation because no human had personally made decisions to discriminate; it was rather a matter of a poorly understood AI model. This argument was dismissed by the Dutch DPA, which insisted that deploying decision-making systems and using their output carries with it a regulatory responsibility.[51] The Dutch DPA's approach was confirmed by the parliamentary report into the matter.[52]

---

[47] FCA Handbook, PRIN 2.1.1. For data protection, see eg General Data Protection Regulation (EU) 2016/679 [2016] OJ L119/1 (GDPR), art 5(1). For further discussion, see Christopher Kuner and others, 'Expanding the Artificial Intelligence-Data Protection Debate' (2018) 8 *International Data Privacy Law* 289.

[48] DSA, arts 16–17, 27, 34–35; OSA, ss 9–12.

[49] Garante per la protezione dei dati personali, Order of 30 March 2023, web doc no 9870847 https://www.garanteprivacy.it/home/docweb/-/docweb-display/docweb/9870847; EDPB, 'Report of the Work Undertaken by the ChatGPT Taskforce' (2024) https://www.edpb.europa.eu/our-work-tools/our-documents/other/report-work-undertaken-chatgpt-taskforce_en.

[50] Commission nationale de l'informatique et des libertés (CNIL), Decision against Clearview AI Inc (16 December 2021, fine issued 20 October 2022), Information Commissioner's Office (ICO), Enforcement Notice against Clearview AI Inc (23 May 2022); Information Commissioner's Office (ICO), Monetary Penalty Notice against Clearview AI Inc (23 May 2022), Garante per la protezione dei dati personali, Decision against Clearview AI Inc (10 February 2022), Hellenic Data Protection Authority (HDPA), Decision No 35/2022 against Clearview AI Inc (13 July 2022).

Required regulatory outcomes for UK financial services firms are set out in the Financial Conduct Authority's (FCA) Principles for Business.[53] It is the responsibility of firms to achieve them, and also the personal responsibility[54] of staff covered by the Senior Managers and Certification Regime (SM&CR).[55] AI policy is being developed slowly. The FCA's published approach to AI in financial services[56] and its consultation on future regulation[57] suggested that it would be unlikely to accept AI autonomy as a defence if a firm fails to achieve the outcomes, but the 2026 Mills Review[58] leaves this to future reform of liability regulation.

At first sight, all this suggests that regulatory outcomes solve the AI autonomy problem. However, the apparently absolute nature of outcome-based regulation is in practice softened greatly once a regulator produces guidance on how regulatees should act. This is because regulators often have enforcement policies which either state[59] or imply[60] that compliance with guidance will limit the regulatee's responsibility to remedying the failure for the future. Additionally, as we explore in section 3.4, the main activity required of regulatees is the design and implementation of suitable governance systems. The best that any governance system can achieve is the mitigation of risk, rather than its complete prevention, and this makes it difficult for regulators to justify imposing stringent penalties for isolated failures.[61] The cumulative effect is to reduce the required level of compliance to what might be described as 'due diligence', a rather higher standard than reasonable care and skill but some way short of an absolute obligation.

---

[51] https://www.autoriteitpersoonsgegevens.nl/documenten/advies-verzamelwet-hersteloperatie-toeslagen.

[52] Tweede Kamer der Staten-Generaal, *Ongekend onrecht* (Unprecedented Injustice) (December 2020), https://www.tweedekamer.nl/sites/default/files/atoms/files/20201217_eindverslag_parlementaire_ondervragingscommissie_kinderopvangtoeslag.pdf. For subsequent investigation and action by the Netherlands DPA see https://www.autoriteitpersoonsgegevens.nl/documenten/onderzoek-belastingdienst-fraude-signalering-voorziening-fsv.

[53] FCA Handbook PRIN 2.1, www.handbook.fca.org.uk/handbook/PRIN/2/1.html.

[54] Under the FCA Senior Management Conduct Rules, breach of which must be reported to the FCA, and which can incur penalties under FCA Handbook DEPP 6, Penalties, https://handbook.fca.org.uk/handbook/depp6.

[55] https://www.fca.org.uk/firms/senior-managers-certification-regime.

[56] UK FCA, AI and the FCA: our approach (2025-6) https://www.fca.org.uk/firms/innovation/ai-approach.

[57] https://web.archive.org/web/20260207015454/https://www.fca.org.uk/publications/calls-input/review-long-term-impact-ai-retail-financial-services-mills-review.

[58] Mills Review, n 24.

[59] UK FCA , 'The Responsibilities of Providers and Distributors for the Fair Treatment of Customers (RPPD)', para 1.3 https://api-handbook.fca.org.uk/files/document/rppd/RPPD_Full_20190329.pdf.

[60] UK ICO, *Regulatory Action Policy*, https://ico.org.uk/media2/about-the-ico/documents/2259467/regulatory-action-policy.pdf.

[61] Thus the UK ICO writes (ibid, p 13):

> … as issues or patterns of issues escalate in frequency or severity then we will use more significant powers in response. This does not mean however that we cannot use our most significant powers immediately in serious or high-risk cases where there is a direct need to protect the public from harm. Our approach will also encourage and reward compliance. Those who self-report, who engage with us to resolve issues and who can demonstrate strong information rights accountability arrangements, can expect us to take these into account when deciding how to respond.

## 3.2 Process and standards-based regulation

The EU AI Act is a clear example of process and standards-based regulation. It regulates developers of AI systems, deployers and users[62] based on a series of classifications: 'high-risk' systems, general purpose AIs, general purpose AIs which pose systemic risks. AI systems which fall outside those categories are largely outside the ambit of the Act. Additionally, the Act sets out a somewhat random list of prohibited uses of AI.[63]

Developers of high-risk systems are required to undergo processes designed to ensure their safety before they are put into use. Once EU standards are developed[64] developers may choose to demonstrate compliance with those standards, in which case conformity with the Act is presumed.[65] Alternatively they may submit the AI system for conformity testing by a national authority.[66]

Developers of general purpose AIs,[67] which includes LLMs, the basis for agentic AI systems, must document the model's training and testing, and provide information to deployers who are adapting the model for particular purposes.[68] These obligations are generic, and so the information provided will not be based on the specific risks of the deployer's implementation. A developer will need to document the known bias risks in training data, but is not required to (and probably cannot) predict how a bank using the model in Frankfurt versus London creates different risk profiles.

If a general purpose AI poses systematic risks, there are additional obligations. The model must be evaluated to identify and mitigate systemic risks, safeguards against misuse must be developed, and the developer must establish monitoring systems and incident reporting.[69] All developers must maintain records and report serious breaches.[70]

Article 26 creates five concrete user obligations: use according to developer instructions, maintaining competent human oversight, ensuring input data quality, monitoring and keeping logs, and suspending the AI if risks emerge.[71] Most of these are obligations of care, implemented via governance, rather than regulatory outcomes to be achieved; the user guarantees that monitoring systems exist, but not that harm will be avoided. The same is true of the user's preparation obligations covering building and monitoring systems, assigning personnel and ensuring data quality; what the user needs to do is context-dependent. Thus, for example, article 26(2) requires appropriate competence and adequate training, but what is appropriate depends on what the system does.[72] However, the obligation to suspend the AI when monitoring shows problems is an absolute one, and article 26(5) requires suspension when users have reason to consider that risks may exist.

One problem with the AI Act's approach to regulation is its detailed specification of governance and oversight processes. Compliance tends to become a binary tick-box activity; was it done, yes or no? A regulator can verify whether testing occurred, whether documentation was completed, whether incidents were reported, without exercising any qualitative judgment. However, whether ticking these boxes actually improves AI safety depends entirely on how well the regulator predicted the list of processes needed to prevent the relevant harms, another ground on which the Act has been criticised.[73]

[62] The Act's terminology is unfortunately different from usage in the AI industry; developers become 'providers', art 3(3), deployers are 'downstream providers', art 3(68), and users are, confusingly, 'deployers', art 3(4) – see also n 1. This section uses the industry usage for consistency across the article.

[63] AI Act, art 5(1).

[64] AI Act, arts 40(2), 41.

[65] AI Act, arts 40(1), 41(3).

[66] AI Act, art 43(1), Annex VII.

[67] AI Act, art 3(63).

[68] AI Act, art 50.

[69] AI Act, arts 51, 55.

[70] AI Act, arts 13, 72.

[71] AI Act, art 26(1)-(6).

[72] AI Act, Recital 104 (human oversight must be genuine and proportionate to the risks of the specific high-risk AI system).

A second problem is that and process and standards-based regulatory framework is simultaneously over-inclusive and under-inclusive. Thus the AI Act is over-inclusive because it locks in rigid process requirements that may not fit the real risks of specific deployments – a requirement that works for credit scoring in banking might be entirely inappropriate for content moderation on platforms. It is under-inclusive because risks which nobody anticipated in 2024 fall through the cracks. Generative AI producing deepfakes does not fit neatly into high-risk categories, for example, so the Act's most demanding obligations do not apply to systems now identified as generating or amplifying harmful content at scale.[74]

Finally, although the AI Act helps regulate AI systems and actors in the AI value chain, it does not fully address harms arising from platform architecture. From the platform-regulation perspective, this matters because online content harms often arise not only from the AI model itself, but from the wider service environment in which outputs are ranked, recommended, moderated and disseminated. Regulation may therefore apply to the AI system and the platform service separately, and so potentially failing to deal with harms created by the way the AI system operates within that platform environment.

## 3.3 Business model challenges

Organisations rarely implement AI just because it is a cool, new technology. They do so because it reduces the costs of current activities, makes them more efficient, and enables new activities. This means that the business model of the organisation is likely to change.

Regulation of activities is always built on a model of how the activity will be carried out by the regulatees. This can create difficulties if that model is embedded in the regulation in such a way that changes to the organisation's business model means the regulation no longer applies, or applies in unintended ways.[75] Autonomous AI has produced this kind of change in the business models of platforms.

In the EU, content made available via platforms is regulated by the Digital Services Act (DSA),[76] and in the UK by the Online Safety Act 2023 (OSA).[77] Both impose obligations on platforms which make third party content available to others, with more stringent obligations on the largest platforms.

---

[73] Fanny Vainionpää, Karin Väyrynen, Arto Lanamaki and Aayush Bhandari, 'A review of challenges and critiques of the European Artificial Intelligence Act (AIA)' (2023) ICIS 2023 Proceedings.
[74] Claire Boine and David Rolnick, 'Why The AI Act Fails to Understand Generative AI' (2025) 26 Minnesota Journal of Law, Science & Technology 61, 65, 93-94, 112-13.
[75] Chris Reed, 'The Law of Unintended Consequences – embedded business models in IT regulation' (2007) 2 JILT http://www2.warwick.ac.uk/fac/soc/law/elj/jilt/2007_2/reed.
[76] Regulation (EU) 2022/2065 of the European Parliament and of the Council of 19 October 2022 on a Single Market For Digital Services and amending Directive 2000/31/EC (Digital Services Act), OJ L 277, 27 October 2022.
[77] For a comparison of the two, see Benjamin Farrand, 'How do we understand online harms? The impact of conceptual divides on regulatory divergence between the Online Safety Act and Digital Services Act' (2024) 16 *Journal of Media Law* 240.

The DSA is built around the assumption that the content it wants to control by means of take-down upon notice, content which is unlawful, will be made available by intermediary services that host, organise, or restrict information provided by users. For the DSA to apply, the information must have been provided by one user and requested by another.[78] However, it is now common for platforms to integrate LLM-based generative AI which allows users to prompt the AI autonomously to generate content which is then posted to the platform. Hacker, Engel and Mauer argue that this content is provided jointly through a collaboration between the user and the platform's AI tools, and thus falls outside the ambit of the DSA.[79] Additionally, where the platform makes content available as an infinite scroll[80] it is hard to conceptualise the receiving user as having requested the content.

The OSA has a wider scope than the DSA, encompassing not only illegal content but also content which is harmful to children. In addition to take-down upon notice, platforms also have duties to monitor content proactively and to implement processes to reduce the risk that illegal and harmful content will be disseminated. It avoids the DSA's business model embedding by defining the scope of its regulation as applying to any:

> internet service by means of which content that is generated directly on the service by a user of the service, or uploaded to or shared on the service by a user of the service, may be encountered by another user, or other users, of the service.[81]

However, this definition still assumes that generated content will originate from an intentional action by the service user. What, for example, of an agentic AI public relations service, where a company gives the AI the goal of promoting its business and then the AI autonomously determines what content to generate and post? If the AI prompts the platform's own AI to generate content, is that content 'generated directly on the service by a user'? Ofcom's investigation into the X platform illustrates this point. Ofcom had powers to assess whether X had properly assessed and mitigated the risk of users encountering illegal AI-generated imagery on the platform. However, its powers did not extend to investigating the creation of illegal images by the standalone Grok service itself.[82] This confirms the OSA's service-based logic: responsibility is clearer for dissemination through a regulated platform than for standalone autonomous content generation.

Agentic AI is so new that its future capabilities are not foreseeable, and we can see that already the embedded business models of the DSA and OSA are under strain. Their safe-harbour logic, which assumes that the platform is hosting information provided by others and that liability turns largely on knowledge of illegality and response to notice, might well become harder to sustain.[83] The issue is therefore not that platform responsibility disappears, but that the legal basis for it becomes more complex where the platform's own systems help generate or structure the content environment.

## 3.4 Governance

From a regulator's perspective, the regulatee must be the primary responsible actor. This is not because it can control every action of the AI, but because it governs the service environment in which the AI acts on behalf of the regulatee. To a large extent, as sections 3.1 to 3.3 show, responsibility turns on whether the provider identified foreseeable risks, adopted proportionate mitigation, maintained adequate governance systems, and adapted them over time. Thus governance is a critical element of all three regulatory approaches.

---

[78] DSA art 3(g)(iii).
[79] Hacker, Engel and Mauer, n 1, 1118.
[80] https://en.wikipedia.org/wiki/Infinite_scrolling.
[81] OSA s 3(1).
[82]Ofcom, 'Ofcom update: Investigation into X, and scope of the Online Safety Act' (3 February 2026) https://www.ofcom.org.uk/online-safety/illegal-and-harmful-content/investigation-into-x-and-scope-of-the-online-safety-act.
[83] Uta Kohl, 'Toxic Recommender Algorithms: Immunities, Liabilities and the Regulated Self-Regulation of the Digital Services Act and the Online Safety Act' (2024) 16 *Journal of Media Law* 301, 308–13.

Wherever regulation requires the regulatee to establish governance systems for technology, those systems have four tasks to perform:

- Identifying where, and how, the technology generates the risks which the regulation aims to avoid. Regulatees therefore need to establish processes to conduct risk assessments before introducing a new technology. In some cases, the regulation may impose a positive duty to do so, specifying to some extent what that risk assessment should investigate.
- Designing and implementing the technology in a way which prevents identified risks from arising, so far as is possible. This requires systems which ensure testing before the technology is put into operation and modify it in the light of that testing.
- Where risks cannot be prevented completely, then:
    - The regulatee needs processes to evaluate whether those risks are justifiable; and if so
    - It needs to identify and implement mechanisms for risk mitigation, eg by identifying and implementing technical safeguards, introducing human oversight, etc.
- Monitoring the technology's performance to identify failures and shortcomings and, where they are discovered, instituting improvements to the technology to prevent future failures or at least to reduce the risks as much as possible.

All governance systems have a practical weakness in that they depend on regulatees identifying risks, assigning responsibility, and adopting proportionate measures. Identifying the risks which are created by conventional, predictable technologies is hard enough. This is because many of those risks arise out of the context in which the technology is used, and thus the risks change as the context changes. Add to this the uncertainty about how autonomous AI will act, and the task of risk prediction becomes much harder. It seems unlikely that any activity involving autonomous AI can be made entirely risk-free. The same problem exists when trying to predict what processes are needed to mitigate risks. Developers and deployers work in ignorance of what each other has done and will do, and the same is true as between deployers and users. Neither side has both knowledge and control at the same time.[84]

This uncertainty makes the task of risk management more difficult. If the nature of the risk is unknown, and the context in which it might manifest is equally unknown, devising monitoring systems and safeguards is largely a matter of informed guesswork. Regulators are, as yet, unable to provide guidance to assist regulatees – the UK ICO guidance on AI relates only to pre-generative AI systems,[85] and work on autonomous AI guidance is still at the research paper or consultation stage for all the regulators reviewed here.[86] This means that regulatees are reluctant to devise and implement safeguards whose effectiveness is uncertain and which might not meet their regulator's (unknown) expectations. Ofcom's first-year review supports this concern, finding that many providers omitted important governance information and failed to identify clearly who was responsible for the assessment process.[87]

---

[84] Fraser and Suzor, n 21, 106.

[85] UK ICO, *Guidance on AI and data protection* (2023) https://ico.org.uk/for-organisations/uk-gdpr-guidance-and-resources/artificial-intelligence/guidance-on-ai-and-data-protection/; *Explaining decisions made with AI*, https://ico.org.uk/for-organisations/uk-gdpr-guidance-and-resources/artificial-intelligence/explaining-decisions-made-with-artificial-intelligence/.

[86] UK ICO, *Information Commissioner's Office response to the consultation series on generative AI*, https://ico.org.uk/about-the-ico/what-we-do/our-work-on-artificial-intelligence/response-to-the-consultation-series-on-generative-ai/; Mills Review, n 24; Ofcom, *Ofcom's strategic approach to AI*, https://www.ofcom.org.uk/about-ofcom/annual-reports-and-plans/ofcoms-strategic-approach-to-ai.

Nonetheless, many regulatees have a positive obligation to attempt this difficult task. Data protection regulation is the clearest example, requiring the principle of privacy by design to inform the development of new technological systems. Cavoukian's foundational work, which underlies the GDPR's obligation to implement data protection by design and by default,[88] holds that privacy must be built into systems architecturally, not added as a retrospective safeguard.[89] Requirements for design-stage analysis of risks and impact are also built into platform content regulation,[90] where providers must assess how service design, functionalities, algorithms and recommender systems may create or increase content-related risks. The FCA actively reviews the risk assessment processes and controls of regulated firms.[91]

This suggests that in relation to agentic AI, it cannot be sufficient for the regulatee merely to have policies in place for reviewing the decisions made by the AI. Instead, the processing scope of the AI has to be restricted through technical means to avoid the situation where the AI can make autonomous decisions which go beyond the controller's expected parameters. In the data protection context it has been held that deploying an AI without such restrictions not only amounts to a breach of the controller's duty but also creates the circumstances for the harm.[92]

Retrospective monitoring to detect failures is also a fundamental element of technology governance. However, devising ways to correct undesired performance from autonomous AI systems is as yet still a research topic rather than a set of implementable mechanisms and technologies.[93]

## 3.5 Regulatory fragmentation

Most regulation focuses on the AI user, imposing regulatory obligations with which the user needs to comply in spite of the difficulties we have discussed earlier in this section. This creates two problems which derive from the wider systems in which the AI user operates.

First, users may fall under multiple regulatory systems – for example, a regulated financial services firm will also be a data controller, and may also have some responsibilities as a platform because it makes content available. This requires regulators to co-ordinate,[94] which can be particularly difficult where activities cross jurisdictional borders. Obligations to monitor imposed by one regulator can trigger the application of other regulatory systems, transforming the regulator's counterfactual question – should the regulatee have known this? - into documented evidence that the regulatee now knows. That new evidence cascades across regulatory regimes; for example, bias found in FCA mandated monitoring could trigger EU AI Act article 26(5) breaches, GDPR violations and FCA regulatory breaches simultaneously.[95] An EU bank deploying credit scoring is simultaneously a GDPR data controller, an AI Act deployer, subject to MiFID II, the Consumer Credit Directive and equality law,[96] as well as its national financial services regulation. These frameworks are not aligned and ask incompatible questions of the same system.

---

[87]Ofcom, *Year 1 Online Safety Risk Assessments: Report on Standards and Improvements* (4 December 2025) 3, 5, 17–18.

[88] GDPR, art 25.

[89]Ann Cavoukian, *Privacy by Design: The 7 Foundational Principles* (Information and Privacy Commissioner of Ontario 2009).

[90] DSA, arts 34–35; OSA, ss 9–12.

[91] FCA, *Risk assessment processes and controls in firms: our findings* (2025) https://www.fca.org.uk/publications/good-and-poor-practice/risk-assessment-processes-and-controls-firms-our-findings.

[92] See the two recent cases involving enforcement actions by the UK ICO against Snap and by the Hamburg DPA against H&M where autonomous decision-making by AI assisted employee profiling systems led to the unjustified accumulation of health and personal beliefs-related information. ICO, *Enforcement Notice: Snap Inc (My AI Chatbot DPIA)* (October 2023, revised March 2024) https://ico.org.uk/about-the-ico/media-centre/news-and-blogs/2024/05/ico-warns-organisations-must-not-ignore-data-protection-risks-as-it-concludes-snap-my-ai-chatbot-investigation/.

[93] See eg Manish Shukla, 'Adaptive monitoring and real-world evaluation of agentic AI systems' (2026) 6 *AI and Ethics* 129; Niclas Flehmig, Mary Ann Lundteigen and Shen Yin, 'Perspectives on a Reliability Monitoring Framework for Agentic AI Systems' *arXiv preprint arXiv:2511.09178* (2025);
Partha Pratim Ray, 'Reliability-by-Design for Agentic GenAI: Turning the AI Risk Atlas Into an EU AI Act-Ready Assurance Case' *IEEE Reliability Magazine* (2026).

Second and more importantly, many of the risks which users are required to identify and guard against have their origins in other parts of the AI supply chain. Autonomous AI risks are in part shaped by developers, AI providers, and integrators who design, supply, adapt, or connect the AI system,[97] and harm can arise from the interaction of these layers rather than from any one isolated decision.[98] AI developers create agentic AI systems which are general purpose by design; they may have some generic model in mind about how those systems might be used, but are unlikely to consider the specific needs of different industries and sectors of commerce in their design, training and testing. Deployers take these generic AI systems and adapt them for the sectors in which their customers operate, but without a full understanding of the limitations and risks which the developer has identified or the working of any safeguards. Even though deployers have some understanding of their chosen sector, they will still be in ignorance of the particular uses which the regulated user has in mind, or might decide on at a later date. When deciding how to use the AI system, users do so in ignorance of the assumptions made by developer and deployer, the risks they have identified, and how any safeguards work.

Thus in order to discharge its regulatory responsibilities properly, the AI user needs to understand the choices made by the members of the supply chain, and to have access to the knowledge those members have acquired as the AI system was developed and adapted. In theory the user could contract with the deployer to receive this information, and similarly the deployer could require it from the developer. In practice, commercial considerations (in particular confidentiality) will limit the exchange of information, and in any event this approach assumes that those who need information can specify what they want, and those who are to provide it can understand those needs.

Thus an autonomous AI failure for which the user is responsible under regulation might be factually the responsibility of some other supply chain member. Alternatively, the failure might arise from the interaction of choices made across the supply chain, all of which are reasonable on their own terms, and thus for which none are fully responsible in moral terms. Regulation does not take a holistic approach to the activity and the supply chain, instead placing all the risk of failure on the user. The consequence is that the user has regulatory liability for risks they cannot understand or control. That user might still be able to foresee general categories of risk, but particular outputs and effects on third parties may be difficult to predict in advance or to guard against.[99]

[94] Four UK regulators, the Competition and Markets Authority, the Financial Conduct Authority, the Information Commissioner's Office and Ofcom, have established the Digital Regulation Cooperation Forum for this purpose, https://www.drcf.org.uk/.
[95] GDPR, art 5(1)(a); European Data Protection Board, 'Guidelines 04/2022 on Calculation of Administrative Fines' (11 October 2022)
[96] Directive 2014/65/EU, arts 24-25; Directive 2008/48/EC, art 5; Directive 2005/29/EC, arts 5-9; Council Directive 2000/78/EC, arts 2-4.
[97] Hacker, Engel and Mauer, n 1, 1115–16.
[98] Rodríguez de Las Heras Ballell, n 29, 7–8, 20.

This analysis suggests that regulation which focuses solely on users is unlikely to be effective at achieving its aims. Regulation needs to recognise that other supply chain members play a part in risk creation and mitigation, and are the source of knowledge which the user will need to meet its regulatory obligations. Whether to extend regulation to the supply chain is not a simple decision – see section 4.1.2 – but the supply chain might be enrolled in regulation indirectly, eg by issuing guidance on what information users should obtain from suppliers.[100]

# 4 Options for regulators

The analysis in sections 2 and 3 shows, in our view, that current regulatory systems need modification to deal with the issues raised by AI autonomy, and in particular agentic AI. Currently they all embed three assumptions which are challenged by AI autonomy.

First, it is assumed that regulatees are able to assess and guard against the risks created by their own activities, and are also able to do this with respect to the technologies they buy in. However, risks are generated in all parts of the AI supply chain, but knowledge is not necessarily passed on between supply chain members. This means that the primary regulatee, the end user of an AI system, may be unaware of some of the risks its AI system presents, and even if aware will not be in a position to control those risks. Thus a holistic regulatory approach to the supply chain is needed. The actors who contributed to the creation or increase of the risk should also be part of the accountability analysis, and the challenge is how to implement effective accountability. Regulation also aims to provide some ex ante assurance to wider society that the regulated activity will be carried on appropriately, and achieving such assurance engages all parts of the supply chain.

A second regulatory assumption is that decisions are made and carried out by humans. This means that regulation does not need precisely defined performance standards, because human competence is recognisable by other humans. The challenge here is in determining when autonomous AI has fallen short and committed a regulatory breach – in other words, we need to decide what quality of decision-making and performance we should expect from AI systems.

A third assumption is that regulatees can give reasons for their decision-making and explain the causes of performance failure, thus enabling regulatory oversight. Achieving AI explainability is still a work in progress, and this is particularly so for those LLMs which power agentic AI. Regulators will need to rethink their mechanisms for regulatory oversight.

## 4.1 Regulating the supply chain

If some of the risks from autonomous AI are outside the control of the AI user, regulation might usefully attempt to extend its scope to the supply chain members who do have ability to mitigate those risks.

[99]Fraser and Suzor, n 21, 111; Rodríguez de Las Heras Ballell, n 29, 7–8; Philipp Hacker, Atoosa Kasirzadeh and Lilian Edwards, 'AI, Digital Platforms, and the New Systemic Risk' (2025) arXiv preprint arXiv:2509.17878, 5, 34.
[100] This would not oblige suppliers to provide that information, but would make it harder for them to resist doing so in contractual negotiations.

### 4.1.1 Through contracts

Where the purchaser of a service is exposed to possible regulatory liability in respect of the service, it is common to negotiate contractual terms under which the service provider agrees to provide information, and makes promises about performance, which reduce the purchaser's risk to an acceptable level. In theory we could expect to find such terms in contracts to provide AI services to regulated entities, and that over time they will converge on an industry understanding of what is appropriate for both parties. Providers are likely to resist such terms for agentic AI because the main risk, that of unexpected and unwanted autonomous AI actions, is so unpredictable.

Regulators can increase the likelihood of contractual terms by imposing on regulatees an obligation to secure performance outcomes from their service suppliers. The UK FCA does this for IT outsourcing through SYSC 8.18, which requires regulatees (inter alia) to ensure that the service provider:

- Discloses 'any development that may have a material impact on its ability to carry out the outsourced functions effectively and in compliance with applicable laws and regulatory requirements';[101]
- Cooperates with the regulator; and
- Allows access by the service user, its auditors and the regulator to data and premises as needed to ensure regulatory compliance.

These obligations have incentivised outsourcing providers to offer financial services-specific terms, which thus bring them indirectly within the regulatory system.[102] Sectoral regulators such as the FCA are likely to want reassurance on these points,[103] and contractual terms are one route to achieving such reassurance.

### 4.1.2 Extending regulated activity regimes to suppliers

Agentic AI suppliers who are located in the EU, or who supply AI services to EU users, are already regulated by the EU AI Act as discussed in section 3.2. However, the general obligations imposed by the AI Act may not correlate well with sectoral regulation, and the Act specifically recognises this for the financial sector by handing supervisory powers under the Act to financial regulators.[104]

The supply chain problem is also recognised in UK financial services law, which in 2025 introduced the Critical Third Parties (CTP) regime. Under s 312L Financial Services and Markets Act 2000, HM Treasury is given power to designate a service supplier as a CTP if 'a failure in, or disruption to, the provision of those services (either individually or, where more than one service is provided, taken together) could threaten the stability of, or confidence in, the UK financial system.' The aim of the CTP regime is to prevent systemic disruption, and the small number of major AI foundation model suppliers suggests that all of them will become candidates for designation once the use of their AI models in the UK financial sector becomes entrenched.

---

[101] FCA Handbook, SYSC 8.18 (6). See also FCA, FG 16/5 Guidance for firms outsourcing to the 'cloud' and other third-party IT services.

[102] Mark Lewis, 'Outsourcing, new technologies and new technology risks: Current and trending UK regulatory themes, concerns and focuses' (2018) 10 *Journal of Securities Operations & Custody* 145; W Kuan Hon and Christopher Millard, 'Banking in the cloud: Part 2–regulation of cloud as "outsourcing"' (2018) 34 *Computer Law & Security Review* 337.

[103] 'Both regulators and consumers need to understand what an agent did, on what authority, and with what result. At the individual level this can resolve consumer complaints but at the aggregate level, this provides the FCA with a supervisory mechanism to detect systemic patterns.' Mills Review, n 24, 145.

[104] AI Act, recital 158.

Designation as a CTP would subject an AI service provider to obligations to ensure its operational risk management and resilience capabilities, primarily through the kinds of governance mechanism discussed in section 3.4. CTPs will be subject to submission of annual self-assessments, scenario testing and incident management exercises, and an annual review by the regulator.[105]

One obvious problem with the CTP regime, or any other mechanism bringing AI service providers within the boundary of a regulated sector, is that some service providers will be foreign entities and their national governments may well object to imposing the regulation on them.[106] The regulatory guidance on the CTP regime proposes cooperation with the service provider's own regulators,[107] which may well be effective if the state in question regulates the service provider in a similar way. If not, UK regulators will need to approach the matter cautiously.

The main weakness of this approach to regulating AI service providers is that it relies on the regulator to analyse and integrate the information it received from the governance systems of its regulated users and any CTPs, and from that to identify systemic risks and guard against them. This requires a level of experience and expertise which may not be available to the regulator. Arguably, the necessary level of expertise is likely only to be found among the staff of major AI providers, and thus regulation might require them to undertake the analysis rather than the regulator.[108]

There is also a serious problem of multiple, overlapping and contradictory regulation, as each national regulator extends its rules to the AI service provider. This may not be a major issue for financial services, which are strongly coordinated at a global level, but would be in the case of eg online content regulation, where national approaches differ markedly. From a purely national perspective it makes obvious sense to enrol foreign AI service providers in national regulatory systems; from a global perspective, much less so.

## 4.2 Establishing AI decision-making standards

A difficult question, which this section poses rather than answers, is: against what standards should AI decision-making be judged? Where regulation is outcome-based the only judgment needed is whether the outcome has been achieved, and the test will be the same for an AI decision as for a human one. However, if the regulation requires an open-textured[109] standard such as reasonable care or best efforts, the question becomes a hard one.

Human performance of open-textured obligations is judged against the expected performance of other humans. In effect, judges imagine themselves in the position of that human and ask what decision or action they would have taken instead. This is not something judges can do for AI decisions or actions.

---

[105] See Bank of England, PRA, FCA, *The Regulators' approach to the oversight of critical third parties* (2024), https://www.bankofengland.co.uk/-/media/boe/files/prudential-regulation/approach/critical-third-parties-approach-2024.pdf, which explains the regulation in more detail and outlines the proposed oversight processes at pp 25-34. See also Mills Review, n 24, 81-2.
[106] For just one example, see *US Government Response to the UK Consultation 'Growing up in the online world'*, https://uk.usembassy.gov/u-s-government-response-to-the-uk-consultation-growing-up-in-the-online-world/.
[107] *Regulators' approach to the oversight of critical third parties*, n 105, 32-3.
[108] Chris Reed, *AI Fairness and Beyond: law, regulation and technology* (Oxford: Hart 2024), Ch 8.
[109] HLA Hart, *The Concept of Law* (2nd, Oxford University Press: Oxford 1994) 128-36.

Where human and AI performance can be measured objectively, as for example in medical scan analysis,[110] regulators could use those metrics to establish minimum performance standards. To achieve social acceptance of AI the minimum standard is likely to require the AI to match human performance, and over time regulators might expect AI to improve over humans and so raise regulatory standards accordingly.

For most human activities, though, generally agreed performance metrics do not exist. These activities tend to be assessed via peer review; doctors judge doctors, academics judge academics, etc. Judges are the exception as the final arbiters, but they are likely to be assisted by evidence from peers of the human being judged.

This last point suggests that the most workable regulatory performance standard for an AI system should be that it performs at least as well as a human would in the same situation.[111] This allows peer review, or evidence from peers if the reviewer is a judge or other external assessor, and sets a standard which is known to be socially acceptable.[112] In effect it translates agentic AI user responsibility into terms already understood and maintains the existing risk profile for users. Users accept the risk that one of their employees might go rogue, and will have to do the same for autonomous AI.

In the longer term it is likely that at least some usable metrics for autonomous AI performance will be developed. In the interim, the standard suggested above seems at least workable, even though it creates uncertainty for both AI developers and AI users, and it can easily be modified as regulated sectors learn more about the potential scope and challenges of AI autonomy.

## 4.3 Improved oversight for risk identification and correction

At present, regulatory oversight is mainly retrospective, based on information and analysis provided by the regulatee. To be effective in controlling risks, those reviewing that material need expertise, authority and practical capacity to act, and the information they are reviewing needs to be complete enough to identify problems. As we saw in section 2, extracting meaningful information about an autonomous AI's working is challenging, and the expertise to understand it is scarce. There is a real risk that the oversight process becomes formal compliance rather than substantive control.[113]

---

[110] Geert Litjens et al, 'A survey on deep learning in medical image analysis' (2017) 42 *Medical image analysis* 60; Dominik Müller, Inaki Soto-Rey and Frank Kramer, 'Robust chest CT image segmentation of COVID-19 lung infection based on limited data' (2021) 25 *Informatics in medicine unlocked* 100681. Note though that even here, metrics can be misleading if used outside the specific context for which they were devised – AS Panayides et al, 'AI in Medical Imaging Informatics: Current Challenges and Future Directions' (2020) 24 *IEEE Journal of Biomedical and Health Informatics* 1837.

[111] We think this true for the regulatory systems examined here. However, readers should note the possibility of socially acceptable cost-benefit trade-offs if the AI is delivering scarce resources. Thus society might accept a medical diagnosis AI which performs a little less well than the average medical specialist if it substantially decreases the time taken to make a diagnosis because experts in that field are in short supply.

[112] This standard is explicitly set out in the UK Automated Vehicles Act 2024, which governs the approval of automated vehicles for use on the road. Section 2(1) requires the Secretary of State to establish principles for determining that issue, and section 2(2) provides:

> The principles must be framed with a view to securing that (a) authorised automated vehicles will achieve a level of safety equivalent to, or higher than, that of careful and competent human drivers …

[113] Anirban Mukherjee and Hannah Hanwen Chang, 'Agentic AI: Autonomy, Accountability, and the Algorithmic Society' (2025) arXiv preprint arXiv:2502.00289; Muhammad Ajmal and Azmat Islam, 'Who Leads When AI Acts? Authority and Accountability in Agentic Organizations' (2025) 3(12) *Policy Journal of Social Science Review* 629, 631, 640.

To avoid this, oversight will need to become more operational. Autonomous AI requires oversight which focuses less on reviewing isolated outputs after the event and more on making system behaviour observable while the system is operating. This means that human reviewers will need real-time access to logs, testing results and system documentation. If problems with an autonomous AI are detected, there will need to be clear escalation routes and, possibly, even authority to suspend use of the AI system where monitoring reveals serious problems.

Undertaking oversight in this way is only really possible through automation, and the Mills Review has recommended that the UK FCA should develop an AI-enabled agentic supervisory model.[114] In addition to supervising at individual firm level, such a supervisory model should be designed to identify system-wide risks which emerge from shared use of AI technologies:

> Financial services firms increasingly rely on shared models, cloud platforms and data sources. Changes in these systems can affect multiple firms simultaneously, with impacts that may not be visible within any one firm, until they evolve into industry-wide patterns of harm.[115]

# 5 Conclusions

The fundamental bases of most regulatory systems – regulatory outcomes, compliance with guidelines and governance for risk mitigation – are still appropriate tools to regulate the use of autonomous and agentic AI. The AI supply chain will need to be brought within the regulatory systems of its customers, and this will require national and cross-border coordination to avoid overlaps and contradictions. Decision-making standards for AI will need to be established, though these might be as simple as 'as good as a human'. Oversight will need to move to the system level and become real-time rather than retrospective, almost certainly undertaken by supervisory AI systems.

There is, though, a further issue which regulators will need to address. Regulation is imposed to manage risk, and up to now it has been for legislators and regulators to identify the risks and then to impose regulatory obligations to manage them. This has worked well for the risks arising from human decision-making. However, it is by no means clear that regulators have (or can obtain) sufficient knowledge about the risks created by user implementations of autonomous AI to prescribe how they should be managed. Similarly, there will be risks known to supply chain members which are not known to regulators or even users. This tells us that regulation will need to mandate a much higher level of information exchange, and quite possibly will need to rely on regulatees to devise individualised risk mitigations[116] rather than telling their entire sector how to do so.

The biggest change is likely to be felt by those in regulated organisations who are responsible for regulatory compliance. Checklists based on the guidance issued by regulators, designed for auditing the past, will not work here. Increasingly, compliance will require active partnerships between regulators and regulatees, sharing information, evaluating risk and modifying AI systems and processes. This transformation of regulation from a reactive process to an active one will be necessary to meet the challenges of AI autonomy.

[114] Mills Review, n 24, 110-14.
[115] Ibid, 110.
[116] For an example analysis of how this might work see Reed, n 108, Ch 8.